%% file: 14236_tsi.tex
\documentclass[letterpaper, 10pt]{article}

\usepackage{graphicx}
\usepackage{amsmath}
\usepackage{float}
\usepackage{amsfonts}
\usepackage[inner=0.75in, outer=0.75in, top=0.75in, bottom=1in, paperheight=11in, paperwidth=8.5in]{geometry}
\usepackage{epstopdf}
\usepackage{float}
\usepackage{amssymb}
\usepackage[utf8]{inputenc}
\usepackage[english]{babel}
\usepackage{amsthm}
\usepackage{color}
\usepackage{forest}

\usepackage{makecell}
\usepackage{booktabs}
\usepackage{cclicenses}
\usepackage{lscape,array}

\usepackage{microtype}
\usepackage{url}
\usepackage{subfiles}
\usepackage{caption}
\usepackage{subcaption}
\usepackage[skip=10pt plus1pt, indent=40pt]{parskip}
\usepackage{framed}
\usepackage[section]{placeins}

\usepackage{tikz}
\usetikzlibrary{decorations.pathmorphing}
\usetikzlibrary{decorations.markings}
\usetikzlibrary{arrows.meta,bending}
\usetikzlibrary{arrows.meta}
\usepackage{bm}

\usepackage{calrsfs}
\DeclareMathAlphabet{\pazocal}{OMS}{zplm}{m}{n}

\usepackage{adjustbox}

\usepackage{algorithm}
\usepackage{algpseudocode}

\usepackage{standalone}

\usepackage{xcolor}
\usepackage{framed}

\colorlet{shadecolor}{blue!20}

\usepackage[obeyFinal,textwidth=0.55in]{todonotes}
\usepackage[hidelinks]{hyperref}
\usepackage[capitalize]{cleveref}

\theoremstyle{definition}

\begin{document}	

\setcounter{page}{1}
\pagenumbering{gobble}

\title{\bf A Meta-Learning Approach to Population-Based Modelling of Structures}	
\author{G.\ Tsialiamanis, N.\ Dervilis, D.J.\ Wagg, K.\ Worden\\
        Dynamics Research Group, Department of Mechanical Engineering, University of Sheffield \\
        Mappin Street, Sheffield S1 3JD, UK
	   }
	\date{}
    \maketitle

\section*{Abstract}
Machine learning has been widely used in recent years for various disciplines, including in the field of structural dynamics. Solutions to problems like structural identification, are offered by machine learning methods relying only on data acquired from structures and minimal knowledge of the physics of the structure which is modelled. A major problem of such approaches is the frequent lack of structural data. Inspired by the recently-emerging field of population-based structural health monitoring (PBSHM), and the use of transfer learning in this novel field, the current work attempts to create models that are able to transfer knowledge within populations of structures. The approach followed here is \textit{meta-learning}, which is developed with a view to creating neural network models which are able to exploit knowledge from a population of various tasks to perform well in newly-presented tasks, with minimal training and a small number of data samples from the new task. Essentially, the method attempts to perform transfer learning in an automatic manner within the population of tasks. For the purposes of population-based structural modelling, the different tasks refer to different structures. The method is applied here to a population of simulated structures with a view to predicting their responses as a function of some environmental parameters. The meta-learning approach, which is used herein is the \textit{model-agnostic meta-learning} (MAML) approach; it is compared to a traditional data-driven modelling approach, that of \textit{Gaussian processes}, which is a quite effective alternative when few data samples are available for a problem. It is observed that the models trained using meta-learning approaches, are able to outperform conventional machine learning methods regarding inference about structures of the population, for which only a small number of samples are available. Moreover, the models prove to learn part of the physics of the problem, making them more robust than plain machine-learning algorithms. Another advantage of the methods is that the structures do not need to be parametrised in order for the knowledge transfer to be performed.

\textbf{Key words: Structural dynamics, machine learning, meta-learning, population-based modelling, multi-task learning.}

\section*{Introduction}
\label{sec:intro}

Modelling the dynamics of structures has been the objective of many researchers for a very long time. The problem has been dealt with in the traditional modelling scheme of \textit{physics-based} models. Arguably, the most successful such modelling method is the \textit{finite element} (FE) method \cite{bathe2006finite}. The aspect of physics-based models that separates them from other modelling techniques is that they are based on the understanding of the underlying physics of the structure that they model. Based on this understanding, the model formulation is properly selected to build the model, calibrate it and make predictions.

\par

Although such models are quite successful, they are often replaced by \textit{data-driven} models. This second type of model is built exclusively based on data acquired from a structure. A reason for selecting such a model could be the complexity of the problem which needs to be solved, leading to lack of understanding of the underlying physics. Other reasons could be the lack of computational power or the inability to model different scales and domains of physics at the same time using physics-based models. Another reason might be that structures and their environment tend to be so complicated that the existing uncertainties violate the universality of physics-based models, a property that has made them so successful.

\par

A wide range of data-driven models has been developed in the domain of \textit{machine learning} \cite{Bishop2}. Models that \textit{learn} the relationships between input and output quantities, such as \textit{neural networks} \cite{goodfellow2016deep}, have been deployed in a wide range of applications and they have been able to solve quite complex problems, such as image and speech recognition \cite{lecun1995convolutional}, natural language processing \cite{devlin2018bert} and generation of real-looking images \cite{goodfellow2014generative}. The aforementioned tasks are quite complicated, and the development of a traditional approach (equivalent to the physics-based approaches for structural dynamics), may be infeasible.

\par

Machine learning offers a convenient way of solving problems in many domains, including structural dynamics; however it has several drawbacks. One major drawback of machine-learning methods is their dependence on the availability of data. Models, such as neural networks and especially deep neural networks, require large amounts of data to be properly trained. In addition to the need for large datasets, the predictive capabilities of such models are largely dependent on the quality of the available data. This quality could refer to the noise content of the data, but could also refer to the range of conditions that the data cover. Since data-driven models are mainly able to \textit{interpolate}, the training data should cover a wide range of conditions of the structure (environmental and operational), for the model to have satisfactory predictive capabilities.

\par

The current work studies the use of \textit{meta-learning} \cite{hospedales2020meta} for modelling of \textit{populations} of structures. Meta-learning has been formed as the domain of machine-learning algorithms that learn how to learn. Algorithms of the specific domain are focussed on building models which learn how to adapt quite fast (i.e.\ with a few training repetitions), and with a small amount of data. This behaviour is often imposed via the use of information from a population of tasks. Therefore, the framework of the problems that meta-learning is used to deal with aligns with problems that often come up in structural dynamics. Work on the subject has already been performed for the purposes of \textit{population-based structural health monitoring} (PBSHM) \cite{gardner2022population}. A first characteristic of such a framework is that data may be available only from a subset of structures of the population, e.g.\ structures which are studied in a laboratory or structures that have been monitored for a long time. The second characteristic is that for new structures, only a few data samples are available, therefore, the existing knowledge about the available structures should be exploited. Hence, the use of meta-learning for population-based modelling of structures is motivated for the current work and more specifically, the use of \textit{model-agnostic meta-learning} (MAML) \cite{finn2017model}.

\par

The layout of the paper is as follows. In the second section, existing ways of dealing with the problem of lack of data are discussed and a brief overview of a meta-learning method - the \textit{model-agnostic meta-learning} algorithm (MAML) - is given. In the third section, the use of a meta-learning algorithm for modelling of a population of structures is discussed. In the fourth section, an application of the MAML algorithm is on a simulated population of structures is presented. Finally, in the fifth section, conclusions are drawn and future work on the topic is discussed.

\section*{The curse of scarcity}

Lack of data is one of the biggest problems of machine learning. Although neural networks are universal approximators \cite{csaji2001approximation}, their performance depends largely on the available data. Data may not be available, but they might also be expensive to acquire; for example, labelling of images might cost a lot financially and in terms of time. Often, the available data are not sufficient to train models with many trainable parameters - the rule of thumb is that ten training samples are required for every trainable parameter of a neural network. Taking into account that neural network models with millions of trainable parameters exist, the need for huge datasets becomes evident. To deal with cases of insufficient data, many new disciplines have emerged; one of them is \textit{transfer learning} \cite{pan2009survey}. The specific discipline is based on transferring knowledge from existing models to new models. This attempt is made to balance the lack of data when training the new model. Transfer of knowledge is attempted motivated by the fact that the tasks of the two models are similar. Therefore, the attempts are focussed on extracting knowledge from an old model that should be relevant to the new task that a new model is called to perform.

\par

There are several ways of performing knowledge transfer. One intuitive and expository way is via the \textit{convolutional neural network} (CNN) \cite{lecun1995convolutional}. The convolutional layers of CNNs, when trained, form small patterns that the images of the training data have in common. If the task of the model is to classify images, the filters should form into different patterns that separate the two images. In this way, feature extraction is performed for every image and the last part of the CNN, which is often a simple \textit{feedforward neural network} (FNN) \cite{Bishop2}, infers a relationship between the extracted features and the image classes. As a result, a convenient knowledge transfer technique would be to use the filters (early layers) of a trained model as a feature extraction module for the new model and train only the last part of the neural network, the FNN \cite{oquab2014learning}, or allow some recalibration of the filters as well. This approach reduces the trainable parameters of the new model and thus the need for training data. The motivation for such an approach is that common image recognition tasks often require recognition of similar attributes. For example, the convolutional layers of a CNN trained to locate cats in images could be transferred in a new model which shall be trained to locate dogs, since both animals have similar snout characteristics.

\par

In structural dynamics and \textit{structural health monitoring} (SHM) \cite{Farrar}, transfer learning is also quite useful. Data from damaged structures are often unavailable, making the application of machine learning methods for SHM infeasible. As a consequence, alternative solutions to traditional approaches are sought. Transfer learning is one of the solutions for such cases \cite{gardner2022population1}. It is common that data samples from damaged states of newly-deployed structures are unavailable. For this reason, data from damaged structures should be exploited to create models that perform damage identification or localisation in new structures, for which damaged data are not available or scarce. 

\par

Although transfer learning is efficient, many of its applications focus on the transfer of knowledge between only two tasks. A natural extension of this is to attempt knowledge transfer from many tasks to the new task, or to transfer knowledge in between models, while training for many different tasks at the same time. Such approaches are those of \textit{multi-task} learning \cite{ruder2017overview}. The framework of some of these approaches is similar to transfer learning, but for more than one task trained simultaneously, rather than extracting knowledge from a pre-trained model.

\par

Similar to transfer learning, characteristic applications of multi-task learning involve parameter sharing between the initial layers of neural networks. Parameter sharing can be hard \cite{caruana1993multitask}, or soft \cite{duong2015low}. In the case of hard parameter sharing, the parameters of initial layers of the neural network are the same for every task and only the last ones are allowed to differ. In the case of soft sharing, even the parameters of the initial layers are allowed to be different, but some similarity is enforced. The idea behind these models is that similar tasks should have similar parts in the models that are used to perform them. Using such approaches could reduce the need for data separately for every task and exploit the data from the whole population of tasks to enhance the performance of the models. Multi-task learning has also been exploited in engineering. In \cite{bull2022hierarchical}, an application on the analysis of the survival trucks in a \textit{fleet} \cite{bull2022hierarchical} (alternative name for population). 

\par

A recently-emerging field aimed at learning with few samples is \textit{meta-learning}. The aim of meta-learning, as given in \cite{titsias2021information}, is to derive data-efficient algorithms that can rapidly adapt to new tasks. The definition implies that meta-learning algorithms can efficiently learn for a small amount of data, and that they can learn quickly in terms of learning iterations. Various algorithms for meta-learning exist, an interesting family of which are those that are based on the main learning algorithm of neural networks - \textit{gradient descent} \cite{Bishop2, goodfellow2016deep}. Such an algorithm is presented in \cite{andrychowicz2016learning}, where the gradient descent algorithm, which is often used to train neural networks, is replaced by a \textit{recurrent neural network} (RNN), more specifically a \textit{long-short term memory} (LSTM) neural network \cite{hochreiter1997long}. The motivation behind such a replacement is that gradient descent algorithms are general and not specifically-targeted on a family of tasks. As a replacement, a neural network is trained to perform the training for a family of tasks. The neural network is trained for a set of available tasks and then tested on new tasks. The results reveal that the algorithm is indeed able to find an appropriate path in the parameter space towards a point where the loss function is at a minimum (locally or globally).

\par

The meta-learning algorithm which is studied in the current work is the model-agnostic meta-learning algorithm (MAML) \cite{finn2017model}. The MAML procedure is given in Algorithm \ref{alg:MAML}. According to the algortithm, the available data from the set of tasks $\pazocal{T}_{tr}$, are used for training. Tasks $\pazocal{T}_{tr}$ have the form $\pazocal{T}_{tr}=\{\tau_{1}, \tau_{2},... \tau_{n} \}$, where $n$ is the number of available training tasks and each task $\tau_{i}$ has the form $\tau_{i}=\{(\mathbf{x}_{1}, \mathbf{y}_{1}), (\mathbf{x}_{2}, \mathbf{y}_{2}), ... (\mathbf{x}_{m}, \mathbf{y}_{m})\}$, where $m$ is the number of available input-output pairs and each tuple $(\mathbf{x}_{i}, \mathbf{y}_{i})$ comprises training input vectors $\mathbf{x}_{i}$ and target output vectors $\mathbf{y}_{i}$. A loss function $\pazocal{L}$ is used, which is common for all tasks. For example, for regression purposes, the \textit{mean-square error} loss function is used. Moreover, two learning rates $\alpha$ and $\beta$ are needed, for the \textit{inner update} and the \textit{outer update} or \textit{meta update}. Finally, a set of testing tasks $\pazocal{T}_{t}$ is used, for which meta updates are not performed. Essentially, these are the tasks that the model is evaluated on during testing time.

\par

\begin{algorithm}
\caption{Model-agnostic meta-learning (MAML)}\label{alg:MAML}
\begin{algorithmic}
\Require Family of training tasks $\pazocal{T}_{tr}$, testing tasks $\pazocal{T}_{t}$ loss function $\pazocal{L}$, $\alpha$, $\beta$ learning rates, model parameters $\mathbf{\theta}$
\For{each training epoch}
    \For{each task $\tau_{i} \in \pazocal{T}_{tr}$}
        \State Sample a set of data $\pazocal{D}^{tr}_{\tau_{i}}$ for training of the current task $\tau_{i}$
        \State $\mathbf{\theta}'_{i} \gets \mathbf{\theta} - \alpha \nabla \pazocal{L}(f_{\mathbf{\theta}}; \pazocal{D}_{\tau_{i}}^{tr})$
    \EndFor
    \State Sample data for meta-training update from every task $\tau_{i} \in \pazocal{T}_{tr}$ and form the 
    \State meta-training dataset $\pazocal{D}^{m}_{\tau_{i}}$
    \State $\mathbf{\theta} \gets \mathbf{\theta} - \beta \nabla_{\mathbf{\theta}} \sum_{\tau_{i} \in \pazocal{T}_{t}} \pazocal{L}(f_{\mathbf{\theta}'_{i}}; \pazocal{D}^{m}_{\tau_{i}})$
\EndFor
\State \textbf{Testing time}
\State Sample a task of interest $\tau_{j}$ from $\pazocal{T}_{t}$ with available data $\pazocal{D}^{t}_{\tau_{j}}$
\For{each task-specific training step}
    \State $\mathbf{\theta} \gets \mathbf{\theta}^{*} - \alpha \nabla \pazocal{L}(f_{\mathbf{\theta}}; \pazocal{D}^{t}_{\tau_{j}})$
\EndFor
\end{algorithmic}
\end{algorithm}

The algorithm is defined as a two-step update. The first step is the inner update, which is a gradient descent step, resulting in updated model parameters $\mathbf{\theta}_{i}$ for training data $\pazocal{D}^{tr}_{\tau_{i}}$ sampled from a training task $\tau_{i}$. After performing the inner update, the updated values of the model parameters $\mathbf{\theta}'_{i}$ are used to calculate the loss for the outer or meta update. For the sum of the latter losses, the gradients are calculated, but not with respect to the updated values $\mathbf{\theta}_{i}$ as would be expected, but for the values of the model parameters $\mathbf{\theta}$ before the inner update. This strategy forces training towards a point $\mathbf{\theta}^{*}$, from where task-specific model-parameter updates shall result in minimisation of the loss function. More intuitively, by updating the model parameters using gradients with respect to the parameters $\mathbf{\theta}$ instead of $\mathbf{\theta}'_{i}$, the error is backpropagated through the inner updates as well. As a result, information is drawn from the inner updates, making the resulting set of parameters $\mathbf{\theta}^{*}$ not an optimal set of parameters for all the tasks, but a set of parameters from where training updates, similar to the inner-loop updates, will result in a task-specifically optimised model.

\section*{Meta-learning for structural modelling}

The MAML algorithm, as presented in the previous section, fits the needs of a population-based modelling of structures. In the case of PBSHM, data might only be available from a small set of structures. These structures are often monitored for years throughout their lifetime or might even be placed within a laboratory and extensively tested. This situation defines a framework slightly different from the one that meta-learning papers present, such as \cite{titsias2021information, finn2017model, andrychowicz2016learning}. In the aforementioned papers, during training, one has unlimited access to the family of the tasks. During every training epoch (inner and outer loop), a new task $\tau_{i}$ is sampled from the family of tasks $\pazocal{T}$. Such a situation is not common for structural dynamics. The number of structures for which data are available is limited. Consequently, in the current work, the framework that will be tested shall be that of availability of many data samples for a small number of training tasks. A neural network model shall be trained using the MAML algorithm and the error of the model shall be tested on tasks from the population, for which only a few samples are available.

\par

\begin{figure}
    \centering
    \includegraphics[scale=0.45]{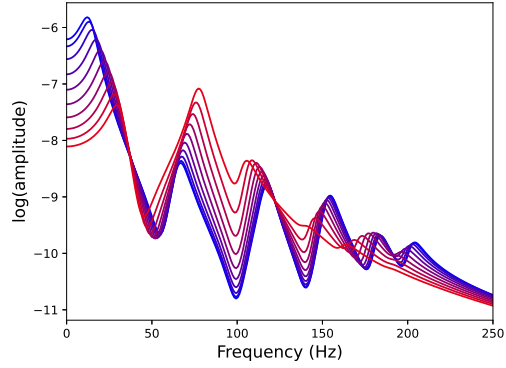}
    \caption{Frequency response functions of the first degree of freedom of a structure of the population for varying temperatures. Lower temperatures correspond to blue curves and as the temperature rises, the transition of the FRF is shown as a gradual change to red colour.}
    \label{fig:varying_temp_FRFs}
\end{figure}

The problem studied here is of a simulated structure, as the one shown in Figure \ref{fig:mass_spring_example}. The population is formed by considering structures with varying stiffness, uniformly sampled from the interval $[8000, 12000]$ and are all considered to be excited by a white noise forcing $F$, applied on the first degree of freedom. The quantity of interest for each structure is the \textit{frequency response function} (FRF), of the first degree of freedom and the varying parameter is the temperature, which is considered to affect the stiffness of the structure according to the nonlinear relationship $k = -13T^{2} + 500T + 7200$. The masses of the structures are both taken equal to unity and the damping parameter $c$ equal to $10$. An example of an FRF for varying temperatures of one of the members of the population is shown in Figure \ref{fig:varying_temp_FRFs}.

\par

\begin{figure}
    \centering
    \includestandalone[width=0.7\textwidth]{Figures/14236_tsi_Fig1}
    \caption{The mass spring system which is used in the current applications, with masses $m$, damping coefficients $c$ and spring stiffness $k$, which is a function of the temperature $T$.}
    \label{fig:mass_spring_example}
\end{figure}

\begin{figure}[H]
\centering
\begin{subfigure}{.5\textwidth}
  \centering
    \includegraphics[width=0.38\paperwidth]{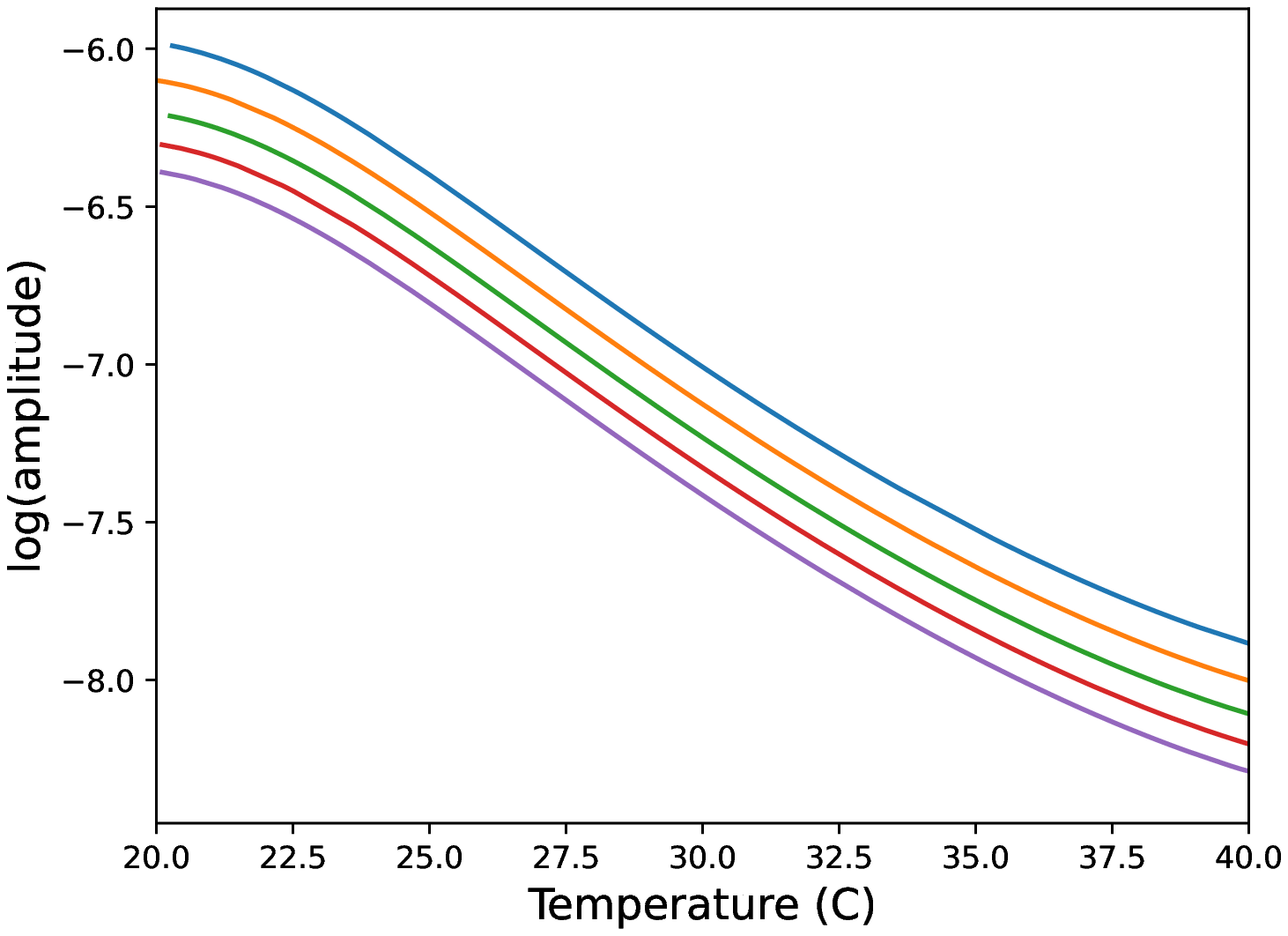}
    \label{fig:1_Hz}
\end{subfigure}%
\begin{subfigure}{.5\textwidth}
  \centering
    \includegraphics[width=0.4\paperwidth]{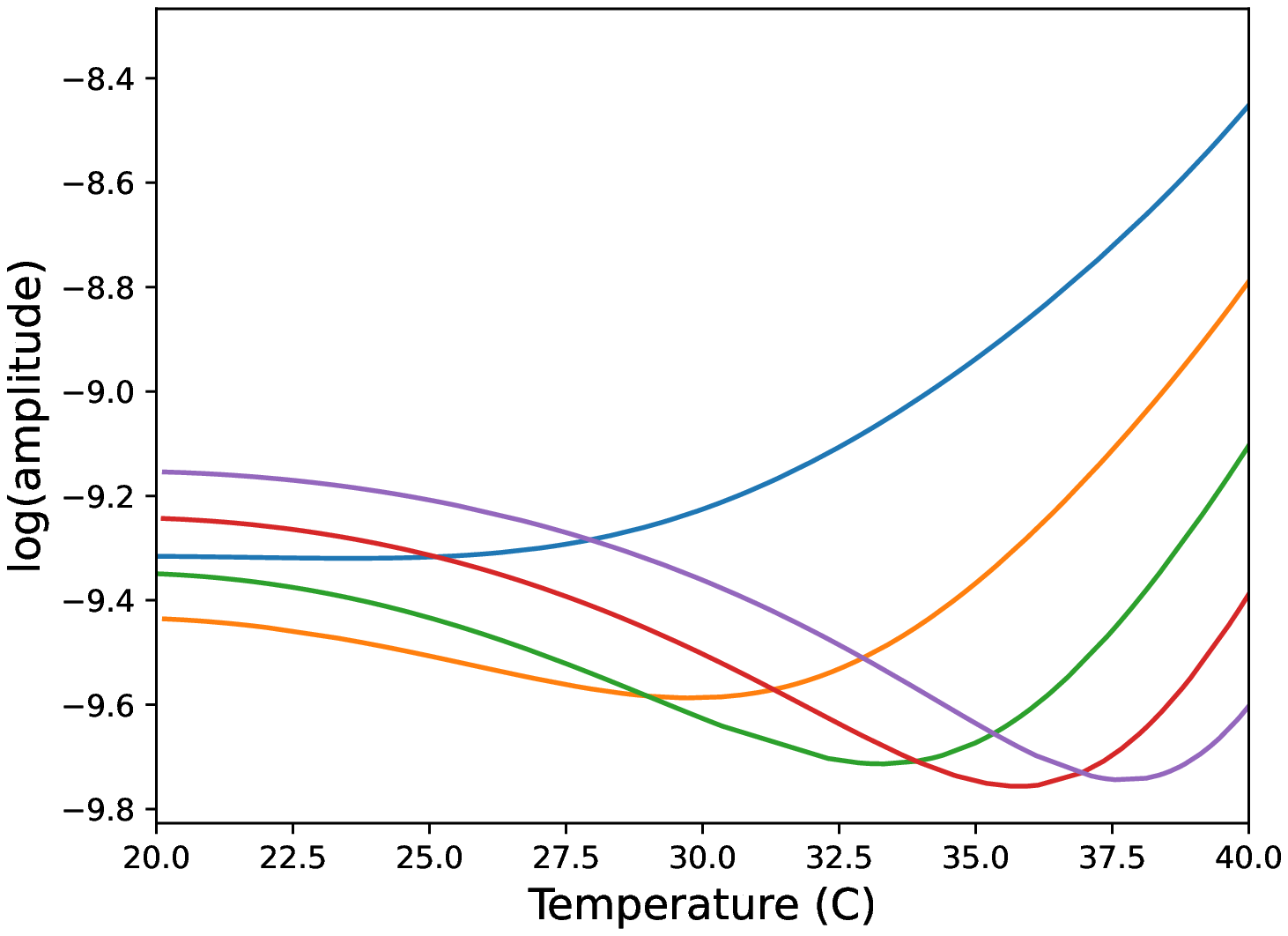}
    \label{fig:100_Hz}
\end{subfigure}
\caption{FRF line corresponding to frequency equal to 1 Hz (left) and 100 Hz (right), for values of the stiffness equal to $8000$ (blue curve), $9000$ (orange curve), $10000$ (green curve), $11000$ (red curve) and $12000$ (curve).}
\label{fig:spectral_lines_example}
\end{figure}

The scenarios considered in the current work were three. The first two scenarios aim at building a model of the population which approximates the value of the magnitude of a single spectral line of the FRF of the first degree of freedom as a function of the temperature. The first scenario referred to the $1$ Hz spectral line and the second on the $50$ Hz spectral line, examples of which are shown in Figure \ref{fig:spectral_lines_example}. 

\par

\begin{figure}
    \centering
    \includegraphics[scale=0.4]{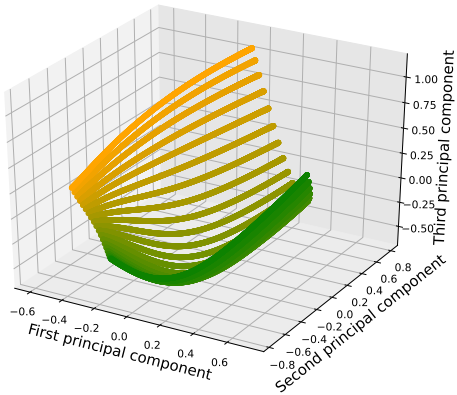}
    \caption{Principal components of samples of FRFs of members of the population. Points belonging to the same structure for different temperatures are shown with the same colour. The colour difference refers to different structures of the population (transition from yellow to green equals transition from a structure with $k=8000$ to a structure with $k=12000$).}
    \label{fig:pca_pop_samples}
\end{figure}

The difference between the two problems becomes clear from the two plots. The first problem is one whose tasks have completely different target values for every value of the input variable (the temperature). For the second problem, overlap between the different tasks is observed. It is expected therefore that the second task shall be more difficult than the first one, since the part of the algorithm's functionality should be the identification of the task, during testing time, in order to properly train the model. The identification in the second case is more difficult because of the overlapping and therefore the performance of the algorithm is expected to be lower in that case. The third problem considered is to approximate the complete FRF. To ensure that the third problem shall be similar to the first (i.e.\ a one-to-one relationship between task and temperature, and the FRF), the FRF of the second degree-of-freedom was also calculated and concatenated with the first to form the target quantities of the dataset. In this way, one shall be more confident that there is no overlap as in the case of the second problem, because by including more characteristics of the structure in the quantities that are inferred, the confidence of a one-to-one relationship between the task and the input is increased. Since the dimensionality of such a quantity of interest is quite large, \textit{principal component analysis} (PCA) \cite{wold1987principal} was performed on the data before applying MAML, to reduce the dimensionality. It was found that the first three principal components explained over $92\%$ of the variance and so they were used as the target values of the algorithm. The three principal components for various members of the population and values of the stiffness are shown in Figure \ref{fig:pca_pop_samples}.

\par

For each of the three problems, different setups were considered for training. The setups are defined by the number of structures available for training - \textit{training structures} - with stiffness values $k$ randomly sampled from the interval $[8000, 12000]$. In order to properly train and test the model, for every scenario, an extra structure with data available was considered, the \textit{validation structure}. The data of the validation structure are used for model-hyperparameter selection. Finally, the model which exhibits the minimum loss function value for the validation structure was tested on the \textit{testing structures}, which in the current work were always $200$ structures, with stiffness values sampled from the same interval as the training and the validation structures. The neural network models used were three-layered neural networks with one-dimensional inputs (the temperature), and outputs according to the number of the modelled quantities, one for the two problems of approximating the magnitude of single spectral lines and three for the case of approximating the principal components of the FRFs. The hyperparameter of the model that is optimised is the size of the hidden layer. 

\par

For every case, the tested hidden-layer sizes were from the set $\{10, 20, ... 100\}$. The models were optimised for number of available training structures being in the set $\{2, 4, 6, 8\}$, and tested for available sample points of the testing structures from the set $\{1, 2, ... 10\}$. The results are presented in terms of the \textit{normalised mean-squared error} (NMSE), given by,
\begin{equation}
    NMSE = \frac{100}{N \sigma_{y}^{2}}\sum_{i=1}^{N}(\hat{y}_{i} - y_{i})^{2}
    \label{eq:NMSE}
\end{equation}
where $\hat{y}_{i}$ is the prediction of the model for the $i$\textsuperscript{th} input sample, $y_{i}$ is the corresponding observation, $\sigma_{y}$ is the standard deviation of the values of the observations $y$ (in the current work, $\sigma_{y}$ refers to the standard deviation of the samples of the whole population) and $N$ is the total number of samples. The NMSE is an objective way to calculate the errors regardless of the scale of the data. Values of the NMSE close to $100\%$ mean that the model provides predictions close to the mean value of the data, while, from experience, values lower than $5\%$ indicate a well-fitted model and values lower than $1\%$ an excellently-fitted model.

\par

\begin{figure}[]
\centering
\begin{subfigure}{.5\textwidth}
  \centering
    \includegraphics[width=0.4\paperwidth]{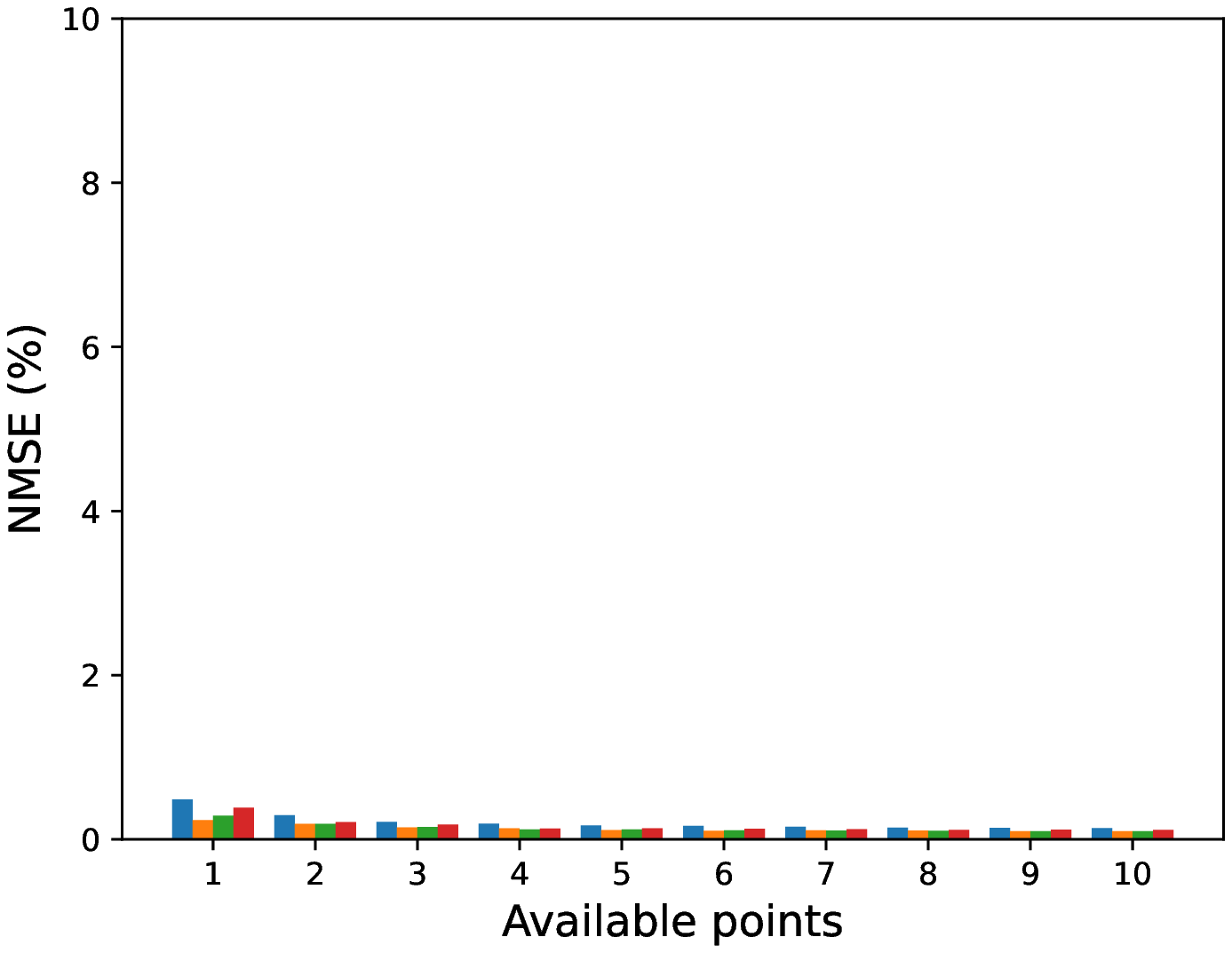}
    \label{fig:1_Hz_erbars_MAML}
\end{subfigure}%
\begin{subfigure}{.5\textwidth}
  \centering
    \includegraphics[width=0.4\paperwidth]{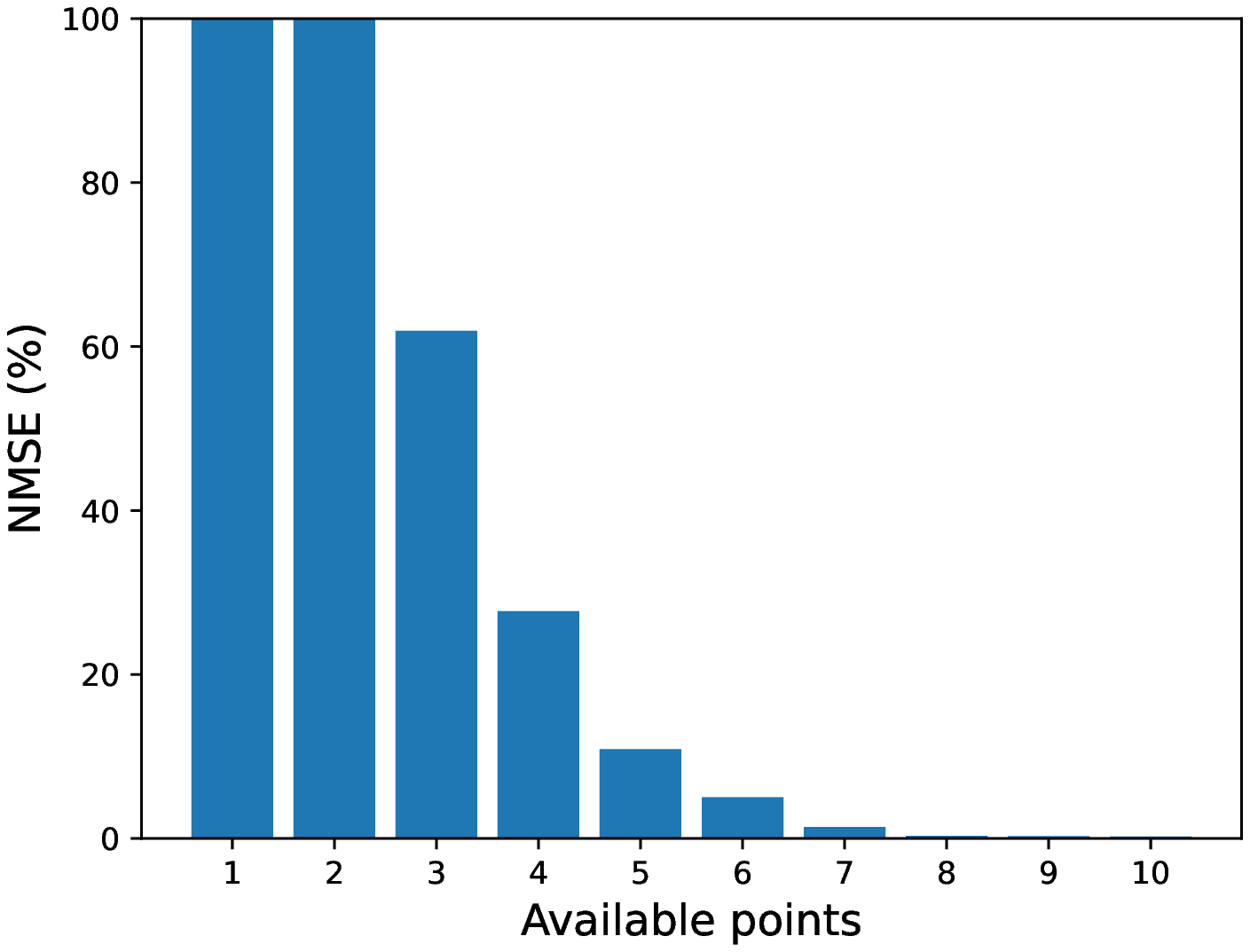}
    \label{fig:1_Hz_erbars_GP}
\end{subfigure}
\caption{Average normalised mean-squared errors for the first problem, for a testing population of $200$ structures and for $100$ data samples for each structure using a neural network trained via the MAML algorithm (left) and for a GP (right). On the left, the different colours represent MAML-trained neural networks with a training population of two (blue), four (orange), six (green) and eight (red) structures.}
\label{fig:error_bars_1_Hz}
\end{figure}

\begin{figure}[]
\centering
\begin{subfigure}{.5\textwidth}
  \centering
    \includegraphics[width=0.4\paperwidth]{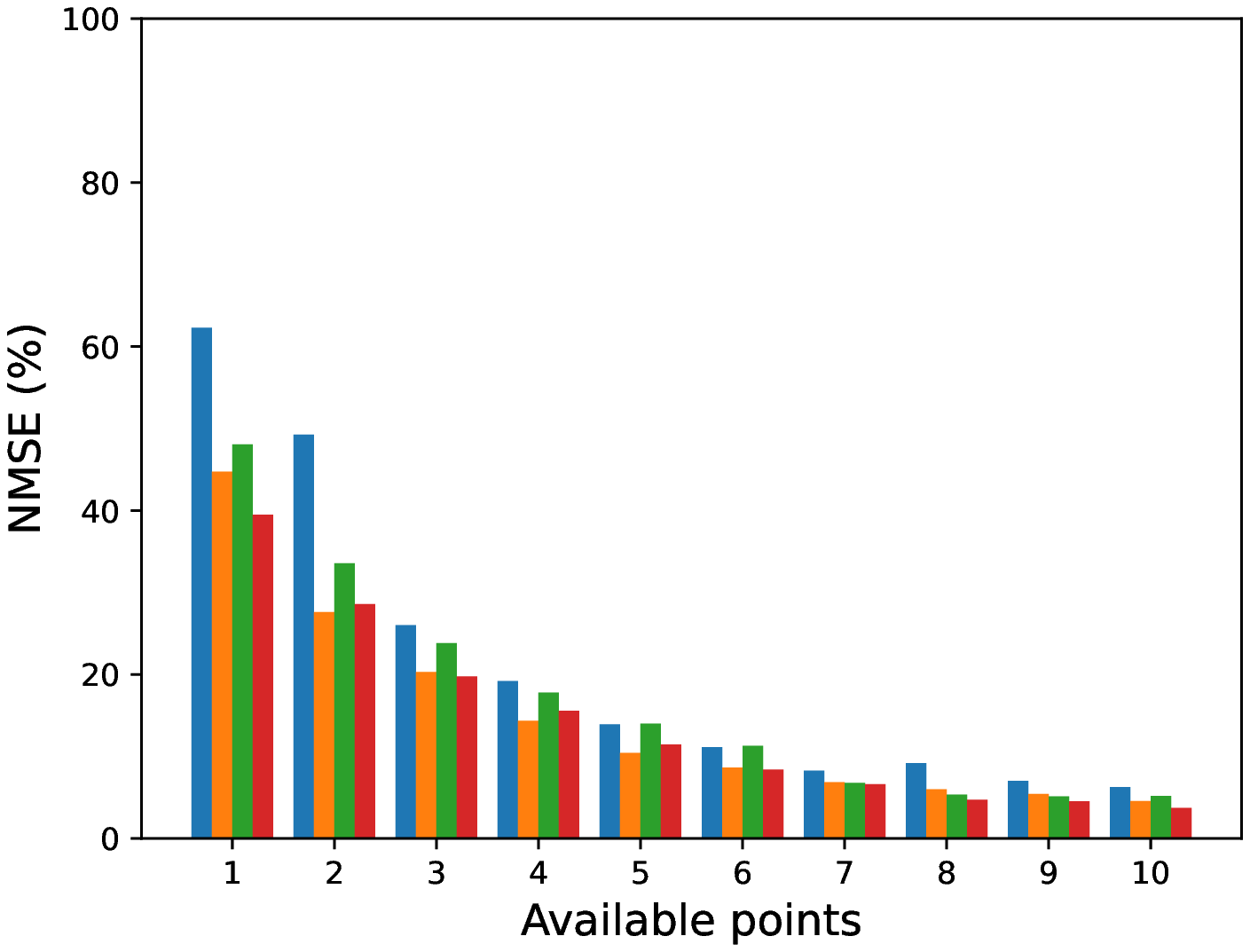}
    \label{fig:50_Hz_erbars_MAML}
\end{subfigure}%
\begin{subfigure}{.5\textwidth}
  \centering
    \includegraphics[width=0.4\paperwidth]{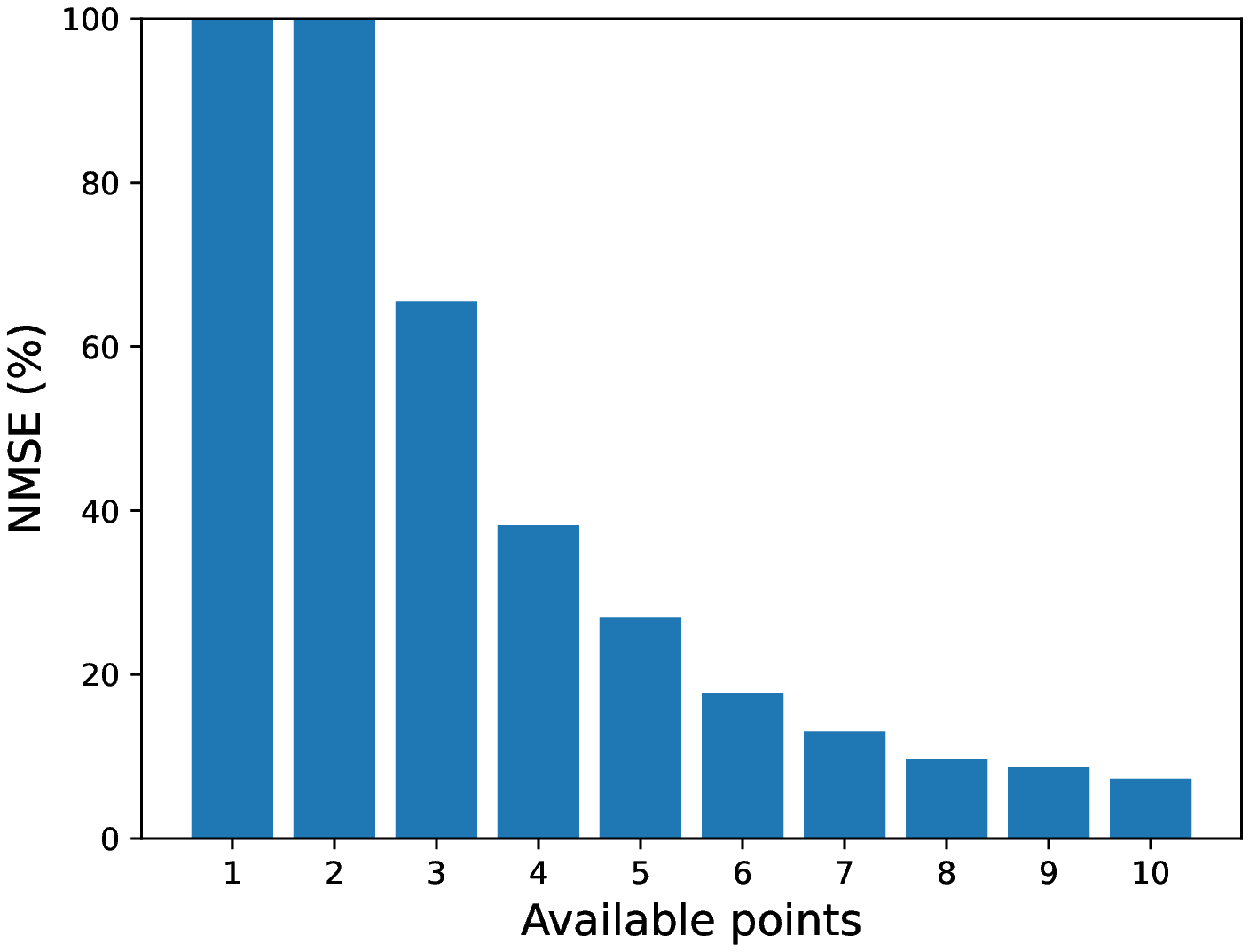}
    \label{fig:50_Hz_erbars_GP}
\end{subfigure}
\caption{Average normalised mean-squared errors for the second problem, for a testing population of $200$ structures and for $100$ data samples for each structure using a neural network trained via the MAML algorithm (left) and for a GP (right). On the left, the different colours represent MAML-trained neural networks with a training population of two (blue), four (orange), six (green) and eight (red) structures.}
\label{fig:error_bars_50_Hz}
\end{figure}

\begin{figure}[]
\centering
\begin{subfigure}{.5\textwidth}
  \centering
    \includegraphics[width=0.4\paperwidth]{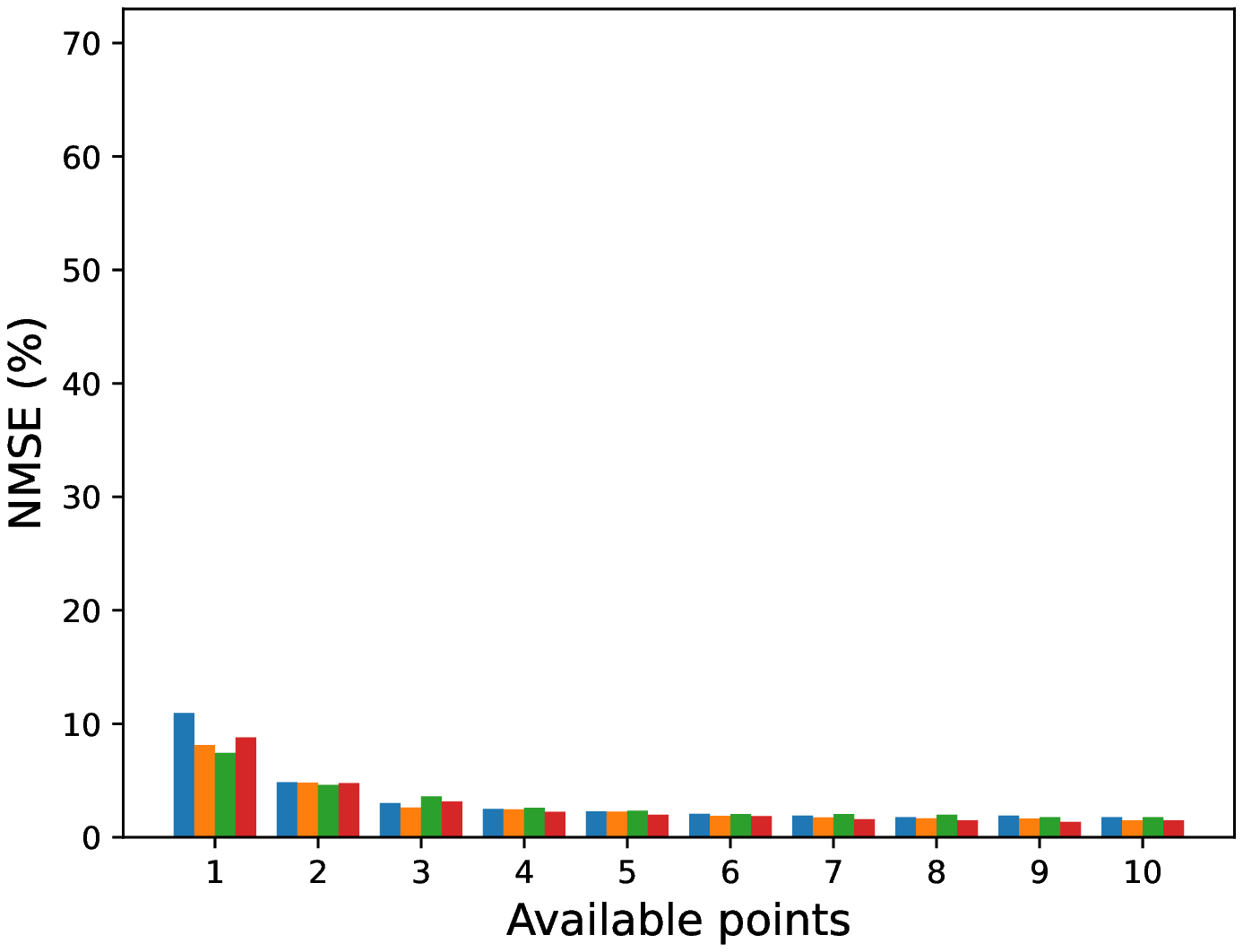}
    \label{fig:FRF_erbars_MAML}
\end{subfigure}%
\begin{subfigure}{.5\textwidth}
  \centering
    \includegraphics[width=0.4\paperwidth]{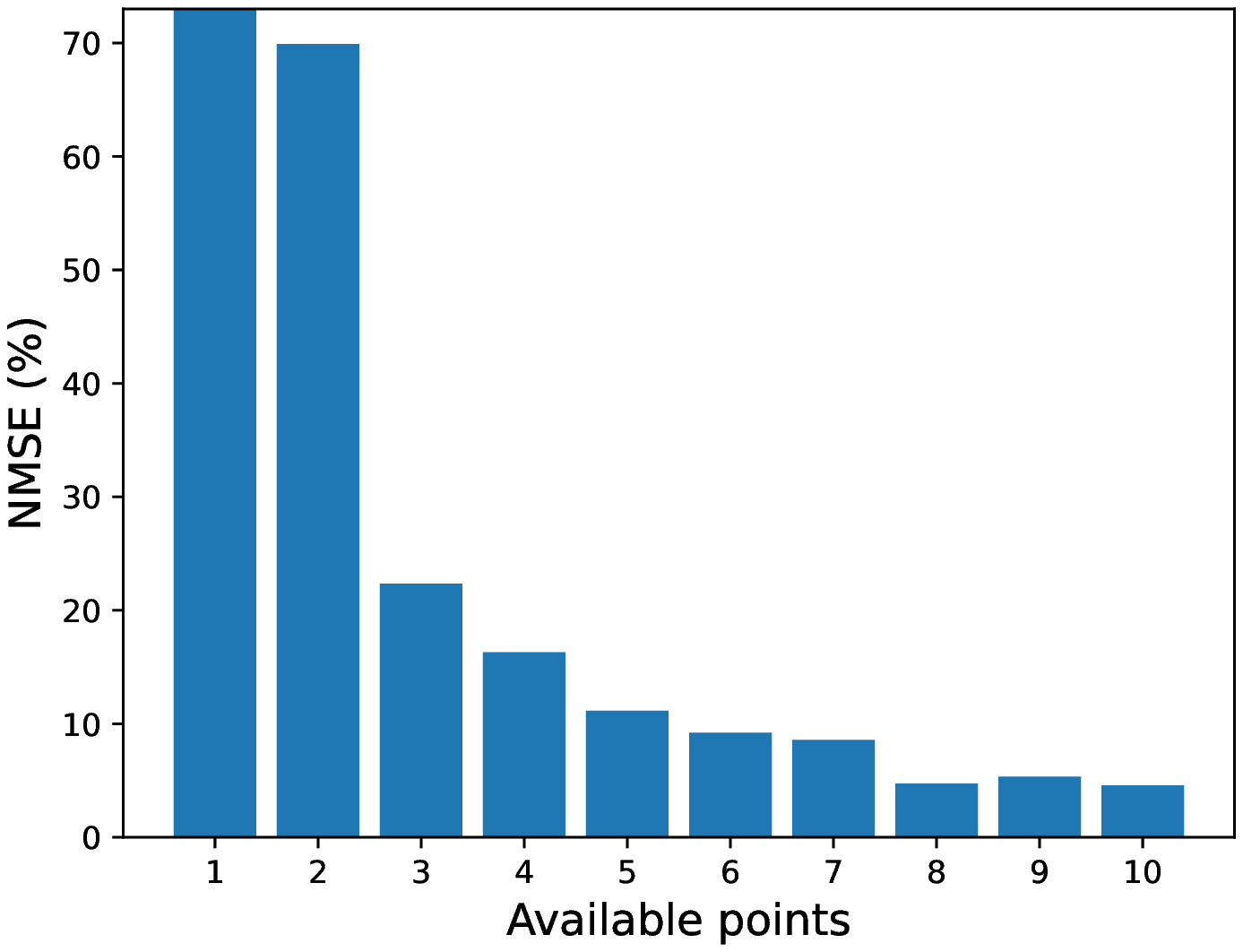}
    \label{fig:FRF_erbars_GP}
\end{subfigure}
\caption{Average normalised mean-squared errors for the third problem, for a testing population of $200$ structures and for $100$ data samples for each structure using a neural network trained via the MAML algorithm (left) and for a GP (right). On the left, the different colours represent MAML-trained neural networks with a training population of two (blue), four (orange), six (green) and eight (red) structures.}
\label{fig:error_bars_FRF}
\end{figure}

The results for the three problems are shown in Figures \ref{fig:error_bars_1_Hz}, \ref{fig:error_bars_50_Hz} and \ref{fig:error_bars_FRF}. On the left-hand side of each figure, the error-bars are shown for various number of available samples for the testing structures during testing time of the MAML algorithm. The errors are the mean NMSEs for $200$ testing structures having $100$ temperature-spectral line couples for each structure. The bars with different colour correspond to different numbers of training structures. On the right-hand side, the corresponding errors are presented for the use of a \textit{Gaussian process} (GP) \cite{rasmussen2003gaussian} with a \textit{radial basis function} (RBF) kernel for $200$ testing structures. The specific algorithm is selected because of its approximation capabilities for small training datasets. 

\par

The results clearly show that the MAML algorithm is able to exploit information acquired from the population. Moreover, it is clear that the increased size of the training population almost always affects positively the errors on the testing population, i.e.\ the higher the number of training structures the lower the error on making predictions for the testing population. Small variations could be from the random selection of the training population for every example. Because the training structures are chosen randomly, in some of the experiments the selection might cover a wider range of values of stiffness than in other experiments, creating a more representative training population. As expected the testing errors are also lower for a higher number of available testing samples. Finally, it is clear that for Problems One and Three, where overlapping between the task curves is not observed, the MAML algorithm performs much better than in the case of Problem Two, where overlapping between the task-specific curves is observed.

\par

It is to the authors' knowledge that the underperformance of the GP is to be expected. The basic formulation of a GP is used and there is no knowledge of the population exploited in order to boost its performance. However, it is also clear that the meta-learning algorithms are able to capture part of the physics of the population without interference of the analyst during training. In contrast, common approaches to introduce such knowledge in the GPs would require prior physical knowledge, i.e.\ defining a \textit{grey-box} model \cite{pitchforth2021grey}. The purpose of the current work is to illustrate how such knowledge can be extracted in an automatic manner. Furthermore, a big advantage of the method is that the tasks do not need to be encoded into vectors; the algorithm is able to identify them (more efficiently in the cases of no overlapping), using exclusively the available data during testing time.

\section*{Conclusions}

In the current work, a population-based meta-learning framework for structural dynamics is presented. The work is focussed on exploiting information structures within a population, to make more accurate predictions for structures of the same population, for which a small number of data samples are available. Meta-learning appears to fit the framework of modelling structures within a population and to exploit information from the data-rich structures to perform inference for data-poor ones. The algorithm used herein was the model-agnostic meta-learning (MAML) algorithm \cite{finn2017model}. The algorithm is developed to fit to new tasks (which in the current case would be new structures) quickly and having only a few samples available. The algorithm is tested on a population of simulated structures. Three problems are considered. Two problems are about approximating the value of the magnitude of a spectral line of the FRF of the structures as a function of the varying temperature, which is considered to affect the stiffness of the structural members, and a third about approximating the whole FRF.

\par

The framework, which is studied here is that of having a set of structures for which many data samples are available - the training structures. Using these structures and the MAML algorithm, neural networks are trained and then tested on many structures of the population, for which a small number of data are considered to be available - the testing structures. The models were trained for various numbers of training structures. The results reveal that the algorithm is indeed able to exploit information from the population and perform inference quite effectively for testing structures with only a few available samples. The algorithm is compared to the use of a Gaussian process, which is an effective algorithm for datasets with a small number of samples. As expected, the MAML algorithm outperforms the GP, which in the current case was not informed in some way by the data of the population. Moreover, the results reveal that MAML is more effective in the case of one-to-one relationships between the task and the controlling variable (the temperature in the current work), and the quantity of interest. It is also clear that further investigation of the effectiveness of the algorithm is needed, and shall be performed in future work, especially regarding the variance of the error in the population and the effect of the number of available training structures.

\section*{Acknowledgements}
\label{sec:ack}

The authors wish to gratefully acknowledge support for this work through grants from the Engineering and Physical Sciences Research Council (EPSRC), UK, via the Programme Grant EP/R006768/. For the purpose of open access, the authors have applied a Creative Commons Attribution (CC BY) licence to any Author Accepted Manuscript version arising.

\bibliographystyle{unsrt}
\bibliography{14236_tsi}

\end{document}

%% file: Figures/14236_tsi_Fig1.tex
    \begin{tikzpicture}
        \draw[line width=0.5mm] (-2,1) -- (-2,-1);

        \draw[line width=0.5mm] (-2, 1) -- (-2.2, 0.8);
        \draw[line width=0.5mm] (-2, 0.8) -- (-2.2, 0.6);
        \draw[line width=0.5mm] (-2, 0.6) -- (-2.2, 0.4);
        \draw[line width=0.5mm] (-2, 0.4) -- (-2.2, 0.2);
        \draw[line width=0.5mm] (-2, 0.2) -- (-2.2, 0.0);
        \draw[line width=0.5mm] (-2, 0.0) -- (-2.2, -.2);
        \draw[line width=0.5mm] (-2, -.2) -- (-2.2, -.4);
        \draw[line width=0.5mm] (-2, -.4) -- (-2.2, -.6);
        \draw[line width=0.5mm] (-2, -.6) -- (-2.2, -.8);
        \draw[line width=0.5mm] (-2, -.8) -- (-2.2, -1.0);

        \node[] at (0.4, 1.4) {$F$};
        \draw[-{>[scale=2.5, length=2, width=3]}, line width=0.5mm] (0.0, 0.5) -- (0.0, 1.0) -- (0.5, 1.0);
        
        \node (1) at (0.0, 0.0) [draw, line width=0.5mm, minimum width=1cm, minimum height=1cm] {$m$};
        \node (2) at (2.5, 0) [draw, line width=0.5mm, minimum width=1cm, minimum height=1cm] {$m$};
        \node (3) at (5.0, 0.0) [draw, line width=0.5mm, minimum width=1cm, minimum height=1cm] {$m$};
        \node (4) at (7.5, 0.0) [draw, line width=0.5mm, minimum width=1cm, minimum height=1cm] {$m$};
        \node (5) at (10.0, 0.0) [draw, line width=0.5mm, minimum width=1cm, minimum height=1cm] {$m$};
        \node (6) at (12.5, 0.0) [draw, line width=0.5mm, minimum width=1cm, minimum height=1cm] {$m$};
        
        \draw[thick, decoration={aspect=0.65, segment length=3mm,
             amplitude=0.2cm, coil}, decorate] (-2, 0) -- (-0.5, 0);
        
        \draw[thick, decoration={aspect=0.65, segment length=3mm,
             amplitude=0.2cm, coil}, decorate] (1) --(2);
        \draw[thick, decoration={aspect=0.65, segment length=3mm,
             amplitude=0.2cm, coil}, decorate] (2) --(3);
        \draw[thick, decoration={aspect=0.65, segment length=3mm,
             amplitude=0.2cm, coil}, decorate] (3) --(4);
        \draw[thick, decoration={aspect=0.65, segment length=3mm,
             amplitude=0.2cm, coil}, decorate] (4) --(5);
        \draw[thick, decoration={aspect=0.65, segment length=3mm,
             amplitude=0.2cm, coil}, decorate] (5) -- (6);

        \node[] at (-1.25, 1.0) {$k(T)$};
        \node[] at (1.25, 1.0) {$k(T)$};
        \node[] at (3.75, 1.0) {$k(T)$};
        \node[] at (6.25, 1.0) {$k(T)$};
        \node[] at (8.75, 1.0) {$k(T)$};
        \node[] at (11.25, 1.0) {$k(T)$};
        
        \draw[line width=0.5mm] (0, -0.5) -- (0, -1.0) -- (-0.5, -1.0);
        \draw[line width=0.5mm] (-0.5, -0.9) -- (-0.5, -1.1);
        \draw[line width=0.5mm]  (-0.4, -1.2) -- (-0.6, -1.2) -- (-0.6, -0.8) --                             (-0.4, -0.8);
        \draw[line width=0.5mm] (-0.6, -1.0) -- (-0.9, -1.0) -- (-0.9, -1.6);
        \draw[line width=0.5mm] (-1.1, -1.6) -- (-.7, -1.6);
        
        \draw[line width=0.5mm] (2.5, -0.5) -- (2.5, -1.0) -- (2.0, -1.0);
        \draw[line width=0.5mm] (2.0, -0.9) -- (2.0, -1.1);
        \draw[line width=0.5mm]  (2.1, -1.2) -- (1.9, -1.2) -- (1.9, -0.8) --                             (2.1, -0.8);
        \draw[line width=0.5mm] (1.9, -1.0) -- (1.6, -1.0) -- (1.6, -1.6);
        \draw[line width=0.5mm] (1.4, -1.6) -- (1.8, -1.6);
        
        \draw[line width=0.5mm] (5, -0.5) -- (5, -1.0) -- (4.5, -1.0);
        \draw[line width=0.5mm] (4.5, -0.9) -- (4.5, -1.1);
        \draw[line width=0.5mm]  (4.6, -1.2) -- (4.4, -1.2) -- (4.4, -0.8) --                             (4.6, -0.8);
        \draw[line width=0.5mm] (4.4, -1.0) -- (4.1, -1.0) -- (4.1, -1.6);
        \draw[line width=0.5mm] (3.9, -1.6) -- (4.3, -1.6);
        
        \draw[line width=0.5mm] (7.5, -0.5) -- (7.5, -1.0) -- (7.0, -1.0);
        \draw[line width=0.5mm] (7.0, -0.9) -- (7.0, -1.1);
        \draw[line width=0.5mm]  (7.1, -1.2) -- (6.9, -1.2) -- (6.9, -0.8) --                             (7.1, -0.8);
        \draw[line width=0.5mm] (6.9, -1.0) -- (6.6, -1.0) -- (6.6, -1.6);
        \draw[line width=0.5mm] (6.4, -1.6) -- (6.8, -1.6);
        
        \draw[line width=0.5mm] (10, -0.5) -- (10, -1.0) -- (9.5, -1.0);
        \draw[line width=0.5mm] (9.5, -0.9) -- (9.5, -1.1);
        \draw[line width=0.5mm]  (9.6, -1.2) -- (9.4, -1.2) -- (9.4, -0.8) --                             (9.6, -0.8);
        \draw[line width=0.5mm] (9.4, -1.0) -- (9.1, -1.0) -- (9.1, -1.6);
        \draw[line width=0.5mm] (8.9, -1.6) -- (9.3, -1.6);
        
        \draw[line width=0.5mm] (12.5, -0.5) -- (12.5, -1.0) -- (12.0, -1.0);
        \draw[line width=0.5mm] (12.0, -0.9) -- (12.0, -1.1);
        \draw[line width=0.5mm]  (12.1, -1.2) -- (11.9, -1.2) -- (11.9, -0.8) --                             (12.1, -0.8);
        \draw[line width=0.5mm] (11.9, -1.0) -- (11.6, -1.0) -- (11.6, -1.6);
        \draw[line width=0.5mm] (11.4, -1.6) -- (11.8, -1.6);
        
        \node[] at (-1.25, -1.0) {$c$};
        \node[] at (1.25, -1.0) {$c$};
        \node[] at (3.75, -1.0) {$c$};
        \node[] at (6.25, -1.0) {$c$};
        \node[] at (8.75, -1.0) {$c$};
        \node[] at (11.25, -1.0) {$c$};

    \end{tikzpicture}